# Hierarchical Bayesian Approach for Improving Weights for Solving Multi-Objective Route Optimization Problem

Romit S Beed

*Dept. of Computer Science, St. Xavier's College, Kolkata, West Bengal, India*

rbeed@yahoo.com

Sunita Sarkar

*Dept. of Computer Science and Engineering, Assam University, Assam, India*

sunitasarkar@rediffmail.com

Arindam Roy

Dept. of Computer Science, Assam University, Assam, India

arindam_roy74@rediffmail.com

Durba Bhattacharya

*Dept. of Statistics, St. Xavier's College, Kolkata, West Bengal, India*

durba0904@gmail.com

The weighted sum method is a simple and widely used technique that scalarizes multiple conflicting objectives into a single objective function. It suffers from the problem of determining the appropriate weights corresponding to the objectives. This paper proposes a novel Hierarchical Bayesian model based on multinomial distribution and Dirichlet prior to refine the weights for solving such multi-objective route optimization problems. The model and methodologies revolve around data obtained from a small scale pilot survey. The method aims at improving the existing methods of weight determination in the field of Intelligent Transport Systems as data driven choice of weights through appropriate probabilistic modelling ensures, on an average, much reliable results than non-probabilistic techniques. Application of this model and methodologies to simulated as well as real data sets revealed quite encouraging performances with respect to stabilizing the estimates of weights.

## I. INTRODUCTION

Majority of the real world complexities generally involve optimizing multiple conflicting objectives. Simply obtaining a solution for their least values concurrently for all the objectives does not guarantee correctness; hence a compromise needs to be made. As these optimization formulations involve multiple objectives, the objective function is formulated as a vector and it is treated as a vector optimization or a multi-objective optimization problem (MOOP) [1]. A multi-objective problem involving multiple, conflicting objectives may be combined into a single-objective scalar function. This approach is named as the weighted-sum method. This is an a priori method established on the “linear aggregation of functions” principle. The method is alternatively named as Single Objective Evolutionary Algorithm (SOEA). By definition, the weighted-sum method reduces to a positively weighted convex sum of the objectives, as follows:

$$\text{Min} \sum_{i=1}^{n} w_i f_i(x), where \sum_{i=1}^{n} w_i = 1;\ w_i > 0\ \forall i \quad (1)$$

Minimization of this single-objective function is expected to give an efficient solution for the original multi-objective problem. The process involves scalarizing the conflicting objectives into a single objective function. There are various scalarization techniques which have been proposed in the past.

Accounting for ambiguity is important when there are restrictions on data which leads to inaccurate interpretation about choices, sensitivities and other behavioral characteristics. Bayesian analysis, grounded on Bayes' theorem, is an instrument that assists in this accounting procedure. Although theoretically lucid, it is hard to apply in different real world problems mainly due to the need of refined estimates. However, this problem was resolved through the advancement of mathematical techniques of iterative calculations largely based on Markov chain Monte Carlo (MCMC) methods. Bayes models free researchers from computational constraints by developing more realistic models of user behavior and decision making by integrating a hierarchical model with Bayesian estimates. The several sub-models are orderly or hierarchically integrated using the Bayes theorem which manages the uncertainty aspect, hence the name Bayesian Hierarchical model.

Bayesian Learning is based on the simple logic that one can achieve better choices by including their recent interpretations and beliefs obtained through previous knowledge and experience. Bayesian learning is also effective where frequentist statistics is not implementable. It possesses supplementary features like iterative upgrade of the posterior while analyzing a hypothesis to assess the parameters of a machine learning model. This promising learning technique is based on Bayes’ theorem to obtain the conditional probability of a hypotheses which is in turn based on certain previous knowledge. Majority of the everyday problems does encompass ambiguity and incremental wisdom, therefore making Bayesian learning more applicable to solve such problems. The Bayesian approach incorporates past belief and gradually refines the prior probabilities based on fresh evidence.

Bayesian data analysis is a process of mathematically modelling data and assigning credibility to individual parameters that are steady with the data and with previous experience. Incorporation of prior belief and experience gives Bayesian learning an edge over frequentist statistics. The Bayesian technique offers greater flexibility in system modeling based on available data. It also provides clarity of demonstrating parameter uncertainty which is accurately inferred and there exists no requirement for building sampling distributions from supporting null hypotheses. Frequentist approaches for calculating ambiguity are inconsistent and tough, whereas Bayesian approaches are essentially intended to offer distinct demonstrations of ambiguity. Frequentist techniques are comparatively more cumbersome for building confidence intervals on parameter. Though there are several advantages of the Bayesian approach, there is an inherent need for an adequately greater set of trials or assigning a confidence to the established hypothesis.

Bayesian reasoning ensures the reorganization of credibility over likelihoods while incorporating fresh data. The main objective of Bayesian estimation is to obtain the most reliable parameter values for the model and this estimation offers a total distribution of credibility over the space of parameter values, not simply one “best” value. The crux of Bayesian estimation is to correctly define how ambiguity changes when fresh data are considered. It is seen that at times that the parameters have significant dependencies on one another. This ordering of dependencies among parameters illustrates a hierarchical model. A hierarchical model specifies dependencies among parameters in an ordered manner based on the semantics of the parameters. Considering data from entities within sets is a salient hierarchical model application. A hierarchical model has the flexibility to possess parameters for every entity that define every discrete entity’s characteristics, and the distribution of different parameters inside a set is exhibited by a higher order distribution with its own parameters that define the tendency of the set. The entity level and set level parameters are assessed concurrently. The hierarchical approach is beneficial as it does not merge the entity’s data together thereby preventing reduction or dilution of trends within entities. To summarize, hierarchical models have parameters that implicitly define the data at several levels and link data within and across levels.

# II. LITERATURE SURVEY

Zadeh popularized the weighted sum technique as a classical approach for solving such problems [2]. This method, as the name suggests, scalarizes a set of conflicting objective functions, by pre-multiplying each of the objective function by predefined weights. The ε-constraints method, introduced by Hamines et al. [3], focused on minimizing the most significant objective function fs(x). Another popular scalarizing technique is the goal attainment technique [4] where the goals are stated for individual objective function fs(x) and the process aims to reduce the overall deviation from the goals. The hierarchical approach [5] and the weighted metrics technique [6] are two more techniques for solving such problems. However, the weighted sum method has gained the most popularity among these due to its simplicity. Although much research has been devoted to the development of different algorithms improving the solution set in multi-objective optimization problems using weighted sum technique, to date, a comprehensive model generating the weights using various sources of uncertainties seem to be lacking. The weighted sum (WS) technique, a commonly used scalarizing technique in multi-objective algorithms, has distinctive advantages of greater search efficiency and easier computational capabilities. Nevertheless, it is frequently critiqued for its inability to predict the logic behind the weight selection as well as its incompetence to deal with nonconvex problems.

It was suggested by Steuer [7] that the weights should scientifically determine the decision-maker’s preference for a particular objective. Das and Dennis [8] offered a graphical explanation of this technique to elucidate few of its drawbacks. The delusion between the hypothetical and the realistic interpretation of the weights for the conflicting objectives made the weight selection process quite inefficient. Various approaches have been suggested for weight selection - Yoon and Hwang [9] suggested a ranking technique whereby the objectives are ranked based on their significance. The most important objective received the largest weight with gradual decrease in weights to lesser important objectives. It was quite similar to the categorization technique in which the conflicting objectives were grouped according to their varying degree of importance. Saaty [10] proposed an eigenvalue process for attaining weights, where n.(n−1)/2 pair-wise evaluations were made between these objective functions to generate a comparison matrix; from this matrix the eigenvalues yielded the weights. Wierzbicki [11] proposed a method for generating weights, where the comparative significance of the objective functions is vague, based on the utopia and the aspiration points. Another method for weight determination was proposed based on fuzzy set theory by Rao and Roy [12]. Though various techniques exist for weight determination, just the selection of the weights may not necessarily generate a feasible solution. New weights may have to be considered and the process may have to be executed again. It was thus suggested by Messac [13] that weights should be functions of the objectives and not simply constants to simulate a task precisely. According to him, the weights must address the issues related to both scaling and relative preference of the objective functions in order to reflect the preference appropriately.

Selection of appropriate weights, lead to an algorithm's better performance. Timothy Ward Athan [14] proposed a quasi-random weighted criteria system that produces weights covering the Pareto set consistently. The method is based on random probability distribution and involves a large number of computations. Gennert and Yuille [15] proposed a nonlinear weight determination algorithm where an optimal point is obtained that is not in the vicinity of the extreme points. Although a lot of work is available in the literature regarding systematic selection of the weights in solving a Multi-Objective Optimization problem, till date a comprehensive data driven technique determining weights reflecting the relative importance of the conflicting objectives is lacking. Many authors including Das and Dennis [8] have shown that choosing weights uniformly over (0,1) does not guarantee uniform spread of Pareto points on the Pareto front. In many cases it has been observed that the points obtained using an uniform generation of weights are found to be clustered in certain regions of the Pareto set. In their subsequent work, Das and Dennis [16] have proposed a technique based on Normal-Boundary Intersection, of obtaining an even spread of Pareto points. Like many others, their method prioritizes the solution set while deciding upon the choice of weights. J. G Lin [17] point towards the scarcity of the number of Pareto Optimal solutions obtained by the existing methods, in addition to some of the solutions coinciding with extreme points. Lin has proposed a method of solving multi-objective optimization problems by transforming them into Single-objective optimization problem, by changing one of the multi-objectives to proper equality constraints using Lagrange multiplier.

Marler and Arora [18] explicated that the weighted sum method is a simple method that delivers a linear estimate of the preference function and need not necessarily reflect one's primary preferences. It is essentially inept of including multifaceted preferences. In spite of determining satisfactory weights a priori, the end solution need not precisely display original preferences. Rui Wang et al. [19] proposed a multi-objective decomposition-based Evolutionary algorithm based on local application of the weighted sum technique. They proposed that the optimal result for each of the search routes is obtained from amongst its adjoining results. Experimental outcomes confirmed that the MOEA/D-LWS outperformed the remaining algorithms for majority of the cases. Zhang proposed a dynamic weighted sum (DWS) technique [20] to methodically alter the weights of individual conflicting objectives for solving multi-objective optimization problems (MOO). He studied the search effect of the different dynamic weighted aggregations namely bang-bang, linear, sinusoidal and random weighted aggregations.

Jaini and Utyuzhnikov [21] proposed a compromise grading method in a fuzzy multi-criteria choice-making system. The fuzzy quantities symbolize the vague weights of each of the conflicting objectives. The authors have designed a fuzzy trade-off grading technique to rank alternatives by awarding the smallest compromise solution as the finest choice. Most of the work in the available literature has focused on fixing the weights based on some prior beliefs or information. The focus of the existing methods is towards refining the distribution of Pareto solutions provided by the WS technique [22], with less emphasis on the stability and appropriateness of the choice of weights for precise representation of the conflicting objectives.

In contrast to the objective of weight determination of the existing works, which aimed at choosing the set of weights which stabilizes the solution set [23-25], this work proposes to frame a model which determines a much stable set of weights in comparison to that obtained deterministically. The criticisms of the existing methodologies for determination of weights have motivated this work and to propose the Bayesian model based on multinomial and Dirichlet priors. As per the authors' existing knowledge, this work is first of its kind since none of the earlier works had this motivation of searching for stability in weights. Unlike the frequentist approach, the Bayesian modelling is based on treating the uncertainties in the parameters probabilistically. The Frequentist methodologies, not taking into account prior probabilities, come up with estimates based mostly on the maximum likelihood or confidence intervals while Bayesians, have a complete posterior distribution over the possible parameter values. This allows them to account for the uncertainty in the estimate by integrating the entire distribution, and not just the most likely value.

This work has been based on the Bayesian framework as one can coherently take into account any prior knowledge (reflected through the prior distribution) about the relative importance of the conflicting objectives to generate the weights. This prior knowledge (distribution) will then be updated using data from the sample using Bayesian paradigm. The sample data obtained from the pilot survey makes the probability distribution narrower around the parameter's true but unknown values. The hierarchical Bayesian model has been so developed as to reflect the relative importance of the conflicting objectives through the respective weights, which were stochastically estimated, based on the data obtained from a pilot survey for the given purpose. Unlike the available continuation techniques, this method can be applied with convenience to handle any number of objectives. As this method yields a posterior probability distribution over the weights, the stochastically generated weight vectors can be used to obtain the points on the Pareto front with less computational complications.

# III. STATISTICAL PREREQUISITE

## BAYESIAN APPROACH

Unlike frequentist approach which does not quantify the uncertainty in fixed but unknown values of the parameters, Bayesian approach, defines probability distributions over possible values of a parameter. Let x denote the data and $\theta$ be the parameter of interest which is unknown. Let $\theta \in \Theta$be the parametric space. Under Bayesian approach one can quantify the prior belief about $\theta$ by defining a prior probability distribution over $\Theta$, the set of possible values of $\theta$ . The newly collected data makes the probability distribution over $\Theta$ narrower by updating the prior distribution to posterior distribution (updated) $\theta$ of using Bayes' theorem which states that

$$P(\theta|\text{Data}) = \frac{P(\text{Data}|\theta)P(\theta)}{P(\text{Data})} \quad (2)$$

where P(Data|θ) is called the likelihood & $P(\theta|Data)$ is the posterior distribution of the parameter θ.

## HIERARCHICAL BAYESIAN MODEL

A model in which the prior distribution of some of the model parameters depend on some other unknown parameters, which are in turn modelled as random variables following some other distribution is a Hierarchical Bayesian model. The level of hierarchy depends on the context and complexity of the problem. Given the observed data x, suppose x follows f(.|θ), with θ being distributed as a prior π(ϕ). If the parameters ϕ can further assumed to be following ζ(υ). The use of hierarchical models ensures a more flexible account of data.

## MULTINOMIAL DISTRIBUTION

It is a multivariate generalization of Binomial distribution. Suppose an experiment is conducted such that each trial has k (finite & fixed) mutually exclusive & exhaustive possible outcomes with probabilities $p_1, p_2, \dots, p_k$ such that $p_i \geq 0\ \forall i = 1(1)k$ and$\sum_{i=1}^{k} p_i = 1$. If $X_i$be the random variable indicating the number of times category I is observed over n independent trials of the experiment, then the vector $\underline{X} = (X_1, X_2, \dots, X_k)$ follows a Multinomial Distribution with parameters n and $p_1, p_2, \dots, p_k$. The probability mass function of the multinomial distribution is:

$$f(X_1 = x_1, X_2 = x_2, \dots, X_k = x_k)$$
$$= \frac{n!}{x_1! \dots x_k!} {p_1}^{x_1} \dots {p_{k-1}}^{x_{k-1}} (1 - p_1 - \cdots p_{k-1})^{\,n - x_1 - \cdots - x_{k-1}} \quad (3)$$

### DIRICHLET DISTRIBUTION

Dirichlet distribution is a multivariate generalization of Beta distribution. Dirichlet distribution of order (k≥ 2) with parameters $\alpha_1, \dots, \alpha_k > 0$, has the following probability density function:

$$g(x_1, x_2, \dots, x_k) = \frac{\Gamma(\sum_{i=1}^{k} \alpha_i)}{\prod_{i=1}^{k} \Gamma(\alpha_i)} \prod_{i=1}^{k} x_i^{\alpha_i - 1} \quad (4)$$

Here Xi's are continuous random variables with $x_i \geq 0 \; \forall i$ and $\sum x_i = 1$, that is, the support of Dirichlet distribution is the set of k-dimensional vectors whose entries belong to (0,1) and add up to one. The parameter vector $p_1, p_2, \dots, p_k$ of the Multinomial distribution has the properties of the $x_i$'s above, as $p=\{p_1,p_2,\cdots,p_k\}$, where $0 \leq p_i \leq 1$ for $i \in [1,k]$ and $\sum p_i = 1$ and hence can be modelled using an appropriate Dirichlet distribution. Dirichlet distribution is a family of continuous probability distribution for a discrete probability distribution with k categories. The usefulness of this method is explained with the help of a realistic example. Considering a company produces six faced dice; though manufacturing processes are precise nowadays, they are still not 100% perfect - if one rolls a randomly selected dice, getting an exact relative frequency of 1/6 for the outcomes is difficult due to a slight manufacturing defect. As one can always expect a probability distribution over the all possible values, 1, 2, 3, 4, 5 and 6; this probability distribution can be modelled using Dirichlet distribution.

### CONJUGATE PRIOR

In Bayesian probability theory if posterior and prior probability distributions of the parameter θ belong to the same probability distribution family, the prior is then called a conjugate prior. In other words, in the formula (2) if P(θ) and P(θ|Data) are in the same family of distributions, they are called conjugate distributions. It can be shown that Dirichlet distribution acts as a conjugate prior for Multinomial distribution.

## IV. PROPOSED METHODOLOGY

As generation of the weights corresponding to different conflicting objectives in a weighted sum problem is the primary interest, the authors have considered the weights $w_i$ in (1) as the unknown parameters.

Let M be a set of conflicting objectives in the objective space defined as follows:

$$\mathcal{M} = \{f_1(x), f_2(x), \dots, f_l(x); h_i(x) \leq / \geq 0, i=1,\dots,p\} \quad (5)$$

where $f_1(x), f_2(x), \dots, f_l(x)$ are the conflicting objective functions, $h_i(x)$ denotes the set of p constraints.

The Weighted Sum method scalarizes the vector objective functions,

$\underline{f} = (f_1(x), f_2(x), \dots, f_l(x)) \epsilon \mathcal{R}^l$, where $\mathrm{R}^l$ is the $l$-dimensional Euclidean space, using the appropriately selected vector of weights $\underline{w} = (w_1, \dots, w_l) \in \mathrm{R}^l$ such that $w_i > 0$ and $\sum_{i=1}^{l} w_i = 1$.

$$\theta = \underline{w}'\underline{f} = w_1 f_1 + \cdots + w_l f_l \qquad (6)$$

It is to be noted that, $\theta \epsilon \mathrm{R}$ is a scalar. Without any loss of generality one can assume the objective functions $f_i(x)\ \forall i = 1(1)l$ to be normalized.

To determine the weights, suppose one obtains data on the preferences of n individuals regarding the choice of different categories (representing different conflicting objectives) through a planned pilot survey. Individuals may be asked to vote for the single most important category out of a finite number of mutually exclusive and exhaustive set of choices. Let $n_i$= number of individuals who have voted for category i (i = 1, 2,...,l; representing the $i^{th}$ objective function) in the pilot survey. The multinomial distribution is used for modeling the probability of counts in the different categories (representing the different objective functions), as the individuals vote independently for exactly one of the l categories. Then, $(n_1, n_2,\ldots,n_l)$ ~ Multinomial $(n; w_1, w_2,\ldots,w_l)$, where $w_i$ is the population proportion of individuals who will vote for category i or is the probability that a randomly selected individual votes for $i^{th}$ category. Probability mass function of multinomial distribution is given by

$$f(n_1,n_2,\ldots,n_l \mid w_1, w_2,\ldots,w_l) =$$

$$\frac{n!}{n_1! n_2! \ldots n_l!}\ w_1^{n_1} w_2^{n_2} \ldots w_{l-1}^{n_{l-1}} (1 - w_1 - \cdots - w_{l-1})^{n - n_1 - \cdots n_{l-1}} \qquad (7)$$

As wi's are continuous random variables, where $w_i \geq 0\ \forall\ i\ and\ \sum w_i = 1$, it can be further assumed that,(w1, w2,…,wl) ~ Dirichlet (α1, α2,…,αl) having the following form of density

$$g\ (w_1, w_2,\ldots, w_l) = \frac{\Gamma(\sum_{i=1}^{l} w_i)}{\prod_{i=1}^{l} \Gamma(w_i)} \prod_{i=1}^{l} w_i^{\alpha_i - 1} \qquad (8)$$

Here, Dirichlet distribution, being a distribution over a probability simplex, is most appropriate for modelling $(w_1, w_2,\ldots,w_l)$. Dirichlet distribution is a multivariate generalization of Beta distribution and acts as a conjugate prior to multinomial where $\alpha_1, \alpha_2,\ldots,\alpha_l$ are the concentration parameters such that $\alpha_i > 0\ \forall$ i= 1 (1) l.

The marginal likelihood function is given by,

$$h\ (n_1,n_2,\ldots,n_l \mid \alpha_1, \alpha_2,\ldots,\alpha_l) =$$

$$\int_{w1,w2,\ldots,wl}^{\cdot} f(n_1, n_2, \ldots, n_l | w_1, w_2, \ldots, w_l) \,.\, g\ (w_1, w_2, \ldots, w_l)\ dw_1 dw_2 \ldots dw_l$$

$$= \frac{\Gamma \sum_{j=1}^{l}(\alpha_j)}{\prod_{j=1}^{l} \Gamma(\alpha_j)} \cdot \frac{n!}{\prod_{j=1}^{l} n_j!} \cdot \frac{\prod_{j=1}^{l} \Gamma(n_j + \alpha_j)}{\Gamma(\sum_{j=1}^{l} \alpha_j + n)} \qquad (9)$$

(9) being the conditional probability of observing the data given $\alpha_1, \alpha_2,\ldots,\alpha_l$, the values of $\alpha_1, \alpha_2,\ldots,\alpha_l$ which maximizes (9) are considered as the estimates.

Now it can be shown that $[w_1, w_2,\ldots,w_l \mid n_1, n_2,\ldots,n_l]$ ~ Dirichlet $(\alpha_1+n_1, \alpha_2+n_2, \ldots, \alpha_l+n_l)$

That is, the posterior distribution of the weights given the data, follows Dirichlet distribution with concentration parameters $(\alpha_1+n_1, \alpha_2+n_2,\ldots,\alpha_l+n_l)$.

Posterior expectations of the weights are given by,

$$W_i^* = E(w_i | n_1, n_2, n_3) = \frac{\alpha_i + n_i}{\sum_{i=1}^{3}(\alpha_i + n_i)}, \ i = 1,2,3 \qquad (10)$$

Hence, estimates of weights can be taken as

$$\hat{W}_i^* = \frac{\hat{\alpha}_i + n_i}{\sum_{i=1}^{3}(\hat{\alpha}_i + n_i)} \qquad (11)$$

where $\hat{\alpha}_1$, $\hat{\alpha}_2$ and $\hat{\alpha}_3$ are the values maximizing (9).

Hence , the objective function gets modified as follows:

$$\text{Minimization of } f = \hat{W}_1{}^*f_1 + \hat{W}_2{}^*f_2 + \hat{W}_3{}^*f_3 \qquad (12)$$

The above modeling technique incorporates the uncertainties in determination of the weights through a Bayesian hierarchical model based on multinomial distribution with Dirichlet prior. As observed in the existing literature $w_i$'s have been estimated simply by the proportion of preference in the respective categories,

$$\hat{W}i = p_i = n_i / n; \ i = 1, 2, \ldots, l; \ n = \sum_{i=1}^{l} n_i \qquad (13)$$

In the following section a comparison is done on the estimates of weights obtained under the two different setups using the error variances and one can see that the proposed Bayesian model clearly out performs (13), especially in case of small sample sizes.

# V. RESULTS & DISCUSSION:

Comparison of estimate of weights has been performed for the Frequentist and Bayesian models. Results obtained from the pilot survey are as follows:

$n_1$=24, $n_2$=11, $n_3$=12, n =47

Under Frequentist setup, the estimated weights are

$$\hat{W}i = n_i / n \qquad (14)$$

As $n_i \sim$ Binomial(n, $p_i$) for i=1, 2, 3, the error variance is given by:

$$V(\hat{W}_i) = Var(n_i / n) = \hat{W}_i(1 - \hat{W}_i)/n \qquad (15)$$

Under Bayesian setup, estimates of weights are given in (11).

Expression for variance with respect to posterior Dirichlet ($\alpha_1$+$n_1$, $\alpha_2$+$n_2$ ,…, $\alpha_l$+$n_l$) distribution:

$$V(\hat{W}_i^*) = \mathrm{Var}\left(\frac{\hat{\alpha}_i + n_i}{\sum_{i=1}^{3}(\hat{\alpha}_i + n_i)}\right) = \frac{n\hat{W}i*(1-\hat{W}i*)}{\{\sum_{i=1}^{3}(\hat{\alpha}_i + n)\}^2} \qquad (16)$$

In order to compare the performance of proposed Bayesian model with the existing frequentist method, one needs to consider samples with varying sizes. The results are shown in Table 1. Although the weights seem to be close, it is clear from the results that the new model outperforms the frequentist one with respect to stability under small sample sizes.

| Sample Size | Input values | | | Frequentist weights | | | Bayesian Weights | | | Error Variance Frequentist | | | Error Variance Bayesian | | | Difference in Error Variance | | | Gain in efficiency |
|---|---|---|---|---|---|---|---|---|---|---|---|---|---|---|---|---|---|---|---|
| Total | $n_1$ | $n_2$ | $n_3$ | $w_1$ | $w_2$ | $w_3$ | w1' | w 2' | w 3' | evd 1 | ev d2 | ev d3 | evs 1 | evs2 | evs3 | $d_1$ | $d_2$ | $d_3$ | |
| 10 | 2 | 3 | 5 | 0.2000 | 0.3000 | 0.5000 | 0.2012 | 0.3003 | 0.4986 | 0.01600 | 0.02100 | 0.02500 | 0.00005 | 0.00007 | 0.00008 | 0.01594 | 0.02092 | 0.02491 | 0.99659 |
| 19 | 4 | 6 | 9 | 0.2105 | 0.3158 | 0.4737 | 0.2116 | 0.3159 | 0.4725 | 0.00870 | 0.01140 | 0.01310 | 0.00010 | 0.00013 | 0.00016 | 0.00859 | 0.01126 | 0.01293 | 0.98763 |
| 21 | 7 | 7 | 7 | 0.3333 | 0.3333 | 0.3333 | 0.3333 | 0.3333 | 0.3333 | 0.01060 | 0.01060 | 0.01060 | 0.00015 | 0.00015 | 0.00015 | 0.01044 | 0.01044 | 0.01044 | 0.98505 |
| 75 | 17 | 25 | 33 | 0.2267 | 0.3333 | 0.4400 | 0.2276 | 0.3333 | 0.4391 | 0.00230 | 0.00300 | 0.00330 | 0.00044 | 0.00056 | 0.00062 | 0.00185 | 0.00243 | 0.00267 | 0.80539 |
| 80 | 20 | 40 | 20 | 0.2500 | 0.5000 | 0.2500 | 0.2507 | 0.4986 | 0.2507 | 0.00230 | 0.00310 | 0.00230 | 0.00051 | 0.00067 | 0.00051 | 0.00178 | 0.00242 | 0.00178 | 0.77817 |
| 100 | 20 | 30 | 50 | 0.2000 | 0.3000 | 0.5000 | 0.2038 | 0.2965 | 0.4997 | 0.00160 | 0.00210 | 0.00250 | 0.00055 | 0.00070 | 0.00084 | 0.00104 | 0.00139 | 0.00165 | 0.65568 |
| 100 | 23 | 48 | 29 | 0.2300 | 0.4800 | 0.2900 | 0.2309 | 0.4787 | 0.2904 | 0.00180 | 0.00250 | 0.00210 | 0.00060 | 0.00084 | 0.00069 | 0.00119 | 0.00165 | 0.00140 | 0.66505 |
| 143 | 39 | 45 | 59 | 0.2727 | 0.3147 | 0.4126 | 0.2731 | 0.3151 | 0.4118 | 0.00140 | 0.00150 | 0.00170 | 0.00100 | 0.00100 | 0.00120 | 0.00040 | 0.00050 | 0.00050 | 0.28571 |
| 167 | 20 | 90 | 57 | 0.1198 | 0.5389 | 0.3413 | 0.1213 | 0.5388 | 0.3399 | 0.00060 | 0.00150 | 0.00130 | 0.00060 | 0.00140 | 0.00130 | 0.00000 | 0.00010 | 0.00000 | 0.00000 |
| 168 | 55 | 56 | 57 | 0.3274 | 0.3333 | 0.3393 | 0.3274 | 0.3333 | 0.3393 | 0.00130 | 0.00130 | 0.00130 | 0.00130 | 0.00130 | 0.00130 | 0.00000 | 0.00000 | 0.00000 | 0.00000 |
| 187 | 47 | 78 | 62 | 0.2513 | 0.4171 | 0.3316 | 0.2526 | 0.4158 | 0.3316 | 0.00100 | 0.00130 | 0.00120 | 0.00100 | 0.00130 | 0.00110 | 0.00000 | 0.00000 | 0.00010 | 0.00000 |
| 229 | 84 | 90 | 55 | 0.3668 | 0.3930 | 0.2402 | 0.3664 | 0.3922 | 0.2414 | 0.00100 | 0.00100 | 0.00080 | 0.00100 | 0.00100 | 0.00080 | 0.00000 | 0.00000 | 0.00000 | 0.00000 |

Table 1: Comparison of the error variance under the Frequentist and Bayesian techniques for different sample sizes.

Efficiency of an estimator T2 with respect to T1 is given by

$$E = \frac{V(T_1)}{V(T_2)} \quad (17)$$

Fig. 1 depicts the performance of the two estimators with respect to the relative gain in efficiency for varying sample sizes. Suppose there are two estimators T1 and T2, relative gain in efficiency of T2 with respect to T1 is given by,

$$G = \frac{(V(T1) - V(T2))}{V(T1)} \quad (18)$$

Note that G ≈ 0, indicates that the two estimators are equally efficient. An estimator T2 is more efficient than T1 if V(T2) ≤ V(T1), G > 0. Calculating the relative gain in efficiency in Table I, it is observed that the gain in efficiency due to the proposed method over the existing one is quite high for small sample sizes.

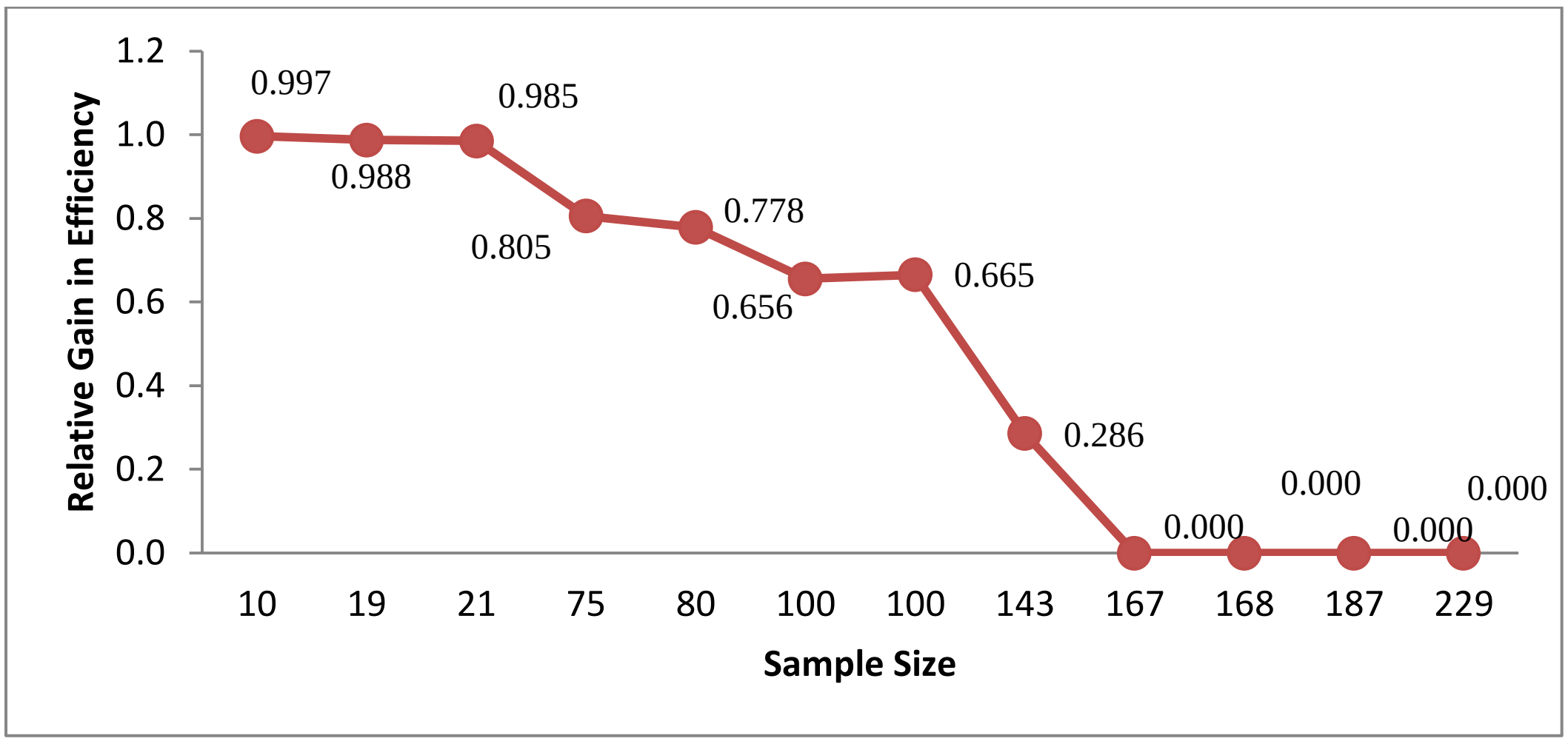

Figure 1: Relative Gain in Efficiency vs Sample Size

It can be seen that the proposed estimator outperforms the existing one with respect to gain in efficiency for small sample sizes. With the increase in sample sizes there is a steady decrease in the gain in efficiency, indicating that with respect to the given data, the two estimators become equally competent for large sample sizes. But in reality, it may be difficult to have a large sample data, thus the effectiveness of the proposed method gains prominence. Bayesian determination of weights is highly recommended in such cases where conducting a large scale survey is time consuming, difficult to implement as well as expensive.

# VI. APPLICATION IN PARKING ROUTE PROBLEM

The proposed methodology has been applied in the field of Intelligent Transport System (ITS). Smart transportation is the need of the hour for sustainable development in a growing economy. Smart transportation supported by a strong communication network and based on sound statistical techniques is a key for tomorrow's smart cities. This route optimization tool promotes environment conservation and sustainable development by providing the most optimal route to a parking lot thereby saving time, energy and fuel. Discovering the most optimal parking lot is a serious problem in the cities and it tends to aggravate during peak hours of the day and at congested places. Selecting the route depends on multiple conflicting objectives, namely minimizing the distance to the parking lot, maximizing the speed of the car and lastly maximizing the parking availability at the lot. The detailed problem definition, formulation, design methodology and implementation are available at [26]. A pilot survey was conducted among 50 drivers. The Bayesian and frequentist weights were calculated respectively using Eqs (11) and (14) and summarized in the table 2.

| Description | Count | Frequentist Weight | Bayesian Weight |
|---|---|---|---|
| Highest priority to distance to parking lot: | 16 | 0.32 | 0.29 |
| Highest priority to travel speed on the route: | 14 | 0.28 | 0.30 |
| Highest priority to parking availability at parking lot: | 20 | 0.40 | 0.41 |

Table 2: Weights - Frequentist and Bayesian

Genetic Algorithm has been used to solve the Multi-Objective optimization problem. The algorithm has been designed to run for 30 generations as the fitness values have stabilized by then in most cases. The fitness values obtained across generations have been plotted and graphically represented in Table 3 and Fig. 2.

| Gen | FREQ 12-4 am | BAYES 12-4 am | FREQ 12 -4 pm | BAYES 12-4 pm | | Gen | FREQ 12-4 am | BAYES 12-4 am | FREQ 12 -4 pm | BAYES 12-4 pm |
|---|---|---|---|---|---|---|---|---|---|---|
| 1 | 0.742 | 0.729 | 0.808 | 0.789 | | 16 | 0.588 | 0.577 | 0.676 | 0.678 |
| 2 | 0.730 | 0.719 | 0.808 | 0.789 | | 17 | 0.588 | 0.566 | 0.676 | 0.654 |
| 3 | 0.711 | 0.696 | 0.808 | 0.750 | | 18 | 0.572 | 0.561 | 0.663 | 0.654 |
| 4 | 0.688 | 0.696 | 0.761 | 0.750 | | 19 | 0.572 | 0.554 | 0.663 | 0.643 |
| 5 | 0.688 | 0.675 | 0.747 | 0.750 | | 20 | 0.572 | 0.550 | 0.663 | 0.632 |
| 6 | 0.671 | 0.672 | 0.718 | 0.732 | | 21 | 0.558 | 0.548 | 0.663 | 0.632 |
| 7 | 0.671 | 0.672 | 0.718 | 0.726 | | 22 | 0.558 | 0.533 | 0.649 | 0.632 |
| 8 | 0.671 | 0.630 | 0.718 | 0.709 | | 23 | 0.558 | 0.533 | 0.649 | 0.615 |
| 9 | 0.654 | 0.630 | 0.702 | 0.709 | | 24 | 0.536 | 0.518 | 0.637 | 0.608 |
| 10 | 0.654 | 0.610 | 0.702 | 0.697 | | 25 | 0.536 | 0.498 | 0.623 | 0.608 |
| 11 | 0.628 | 0.599 | 0.702 | 0.691 | | 26 | 0.529 | 0.498 | 0.623 | 0.596 |
| 12 | 0.628 | 0.599 | 0.695 | 0.678 | | 27 | 0.529 | 0.485 | 0.623 | 0.596 |
| 13 | 0.604 | 0.597 | 0.695 | 0.678 | | 28 | 0.503 | 0.485 | 0.623 | 0.596 |
| 14 | 0.604 | 0.597 | 0.676 | 0.678 | | 29 | 0.503 | 0.485 | 0.623 | 0.596 |
| 15 | 0.588 | 0.596 | 0.676 | 0.678 | | 30 | 0.503 | 0.485 | 0.623 | 0.596 |

Table 3: Fitness values across Generations for Frequentist and Bayesian Weights for two time slots 12AM - 4AM and 12Noon- 4PM

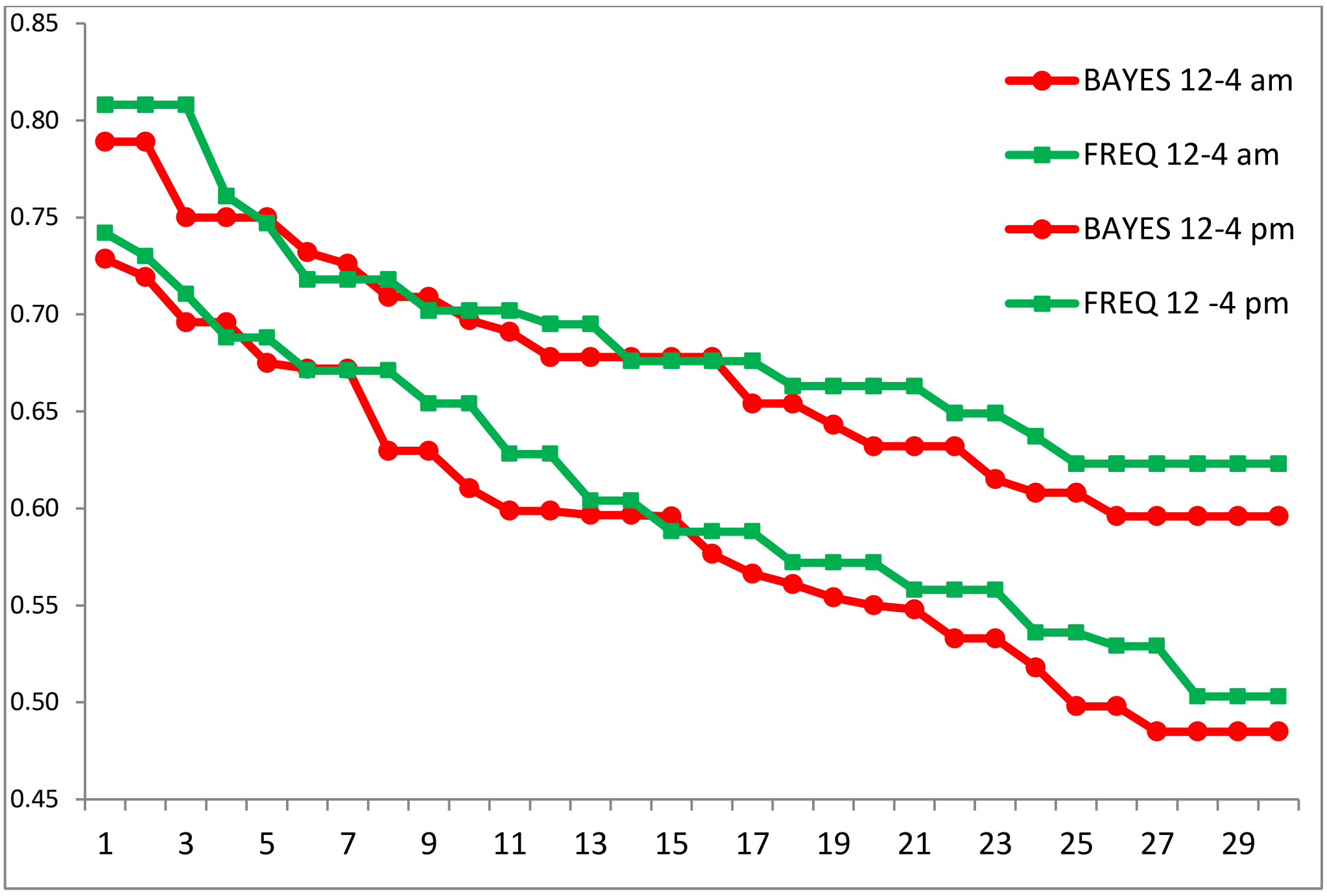

Figure 2: Plot of Fitness vs. Generations for two time slots 12AM - 4AM and 12 Noon - 4PM showing how Fitness values change across 30 generations using frequentist (green) and Bayesian (red) weights.

It is observed that as the generations increases, the value of the fitness function tend to decrease till it stabilizes at an optimal value. In both cases it was seen that the Bayesian weights produced lower fitness values consistently. The process was then repeated for thirty different executions. The fitness values obtained for both the Frequentist and Bayesian weights were noted in Table 4 and plotted in Figure 3.

| RUN | FREQUENTIST FITNESS | BAYESIAN FITNESS | | RUN | FREQUENTIST FITNESS | BAYESIAN FITNESS | | RUN | FREQUENTIST FITNESS | BAYESIAN FITNESS |
|---|---|---|---|---|---|---|---|---|---|---|
| 1 | 0.503 | 0.485 | | 11 | 0.612 | 0.581 | | 21 | 0.360 | 0.349 |
| 2 | 0.622 | 0.589 | | 12 | 0.638 | 0.648 | | 22 | 0.558 | 0.533 |
| 3 | 0.578 | 0.565 | | 13 | 0.476 | 0.413 | | 23 | 0.230 | 0.209 |
| 4 | 0.754 | 0.770 | | 14 | 0.297 | 0.275 | | 24 | 0.287 | 0.295 |
| 5 | 0.688 | 0.659 | | 15 | 0.538 | 0.576 | | 25 | 0.248 | 0.226 |
| 6 | 0.432 | 0.408 | | 16 | 0.368 | 0.329 | | 26 | 0.659 | 0.633 |
| 7 | 0.587 | 0.598 | | 17 | 0.298 | 0.284 | | 27 | 0.710 | 0.699 |
| 8 | 0.389 | 0.342 | | 18 | 0.485 | 0.480 | | 28 | 0.487 | 0.438 |
| 9 | 0.473 | 0.438 | | 19 | 0.422 | 0.439 | | 29 | 0.687 | 0.681 |
| 10 | 0.343 | 0.311 | | 20 | 0.765 | 0.734 | | 30 | 0.522 | 0.498 |

Table 4: Fitness values across 30 executions for Frequentist and Bayesian Weights for time slot 12AM - 4AM

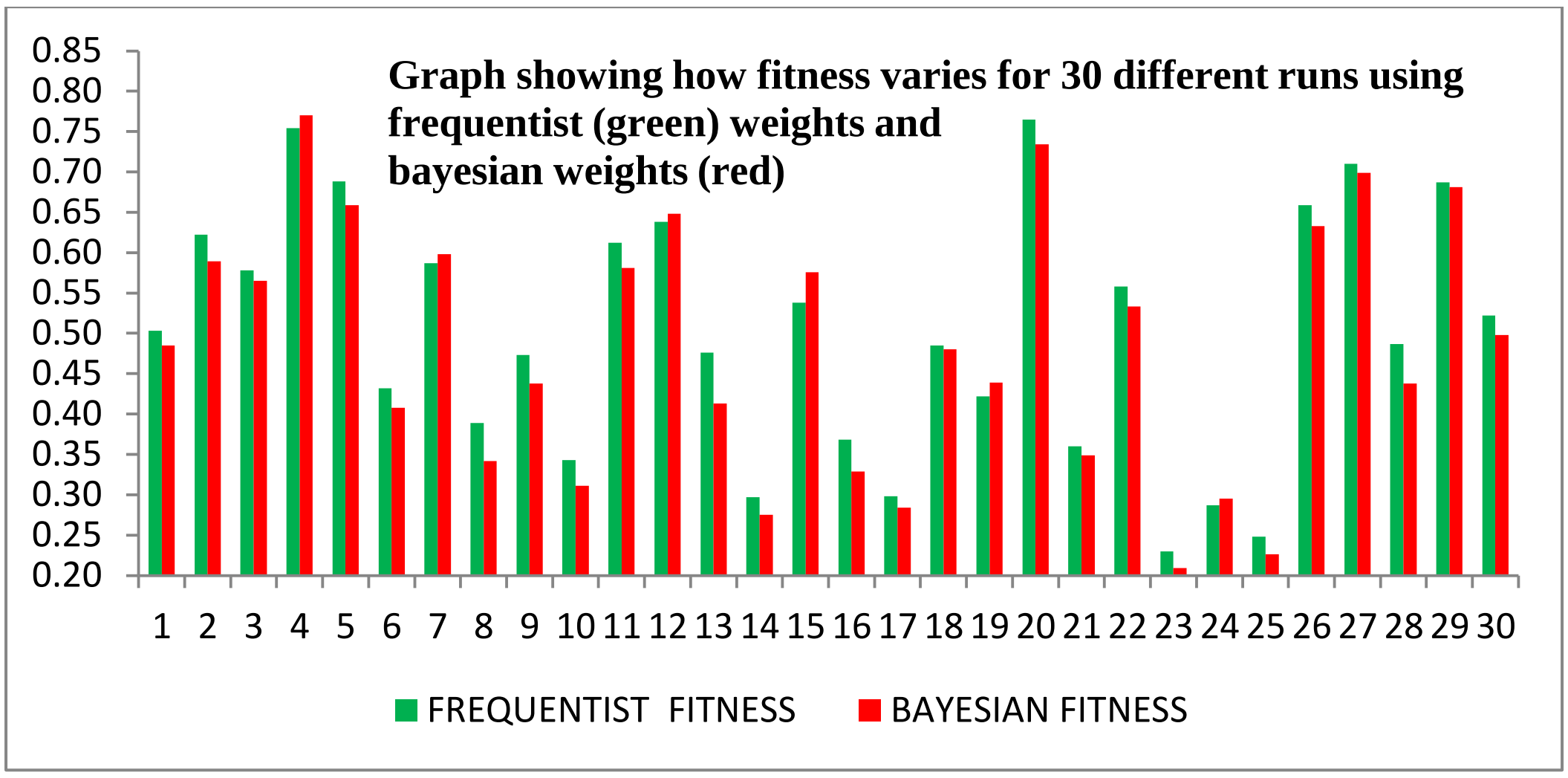

Figure 3: Plot Fitness values across 30 executions for Frequentist and Bayesian Weights for time slot 12AM - 4AM

| Methodology | Mean | Worst | Best |
|---|---|---|---|
| Frequentist Approach | 0.5005 | 0.7650 | 0.2300 |
| Bayesian Approach | 0.4828 | 0.7700 | 0.2090 |

Table 5: Descriptive Statistics for comparing the Fitness values across 30 executions for Frequentist and Bayesian Weights for time slot 12AM - 4AM

Table 3, combined with Figure 2, shows that the fitness value exhibits a consistently decreasing trend as the number of generations increase across all time zones. Focusing on the fitness values corresponding to a typical time zone, here 12 am to 4 am, it can further be noticed from Table 5 that both the average fitness value and the best fitness value were lower for the Bayesian weights than Frequentist weights in 30 executions. Secondly the routes as well as the parking lot vary depending on the time zone. This simulates a real life scenario where parking lots and routes are bound to change as the values for the different factors changes. Although distance remains constant but the average speed and availability of parking lots changes with time which gets finally reflected in the fitness function.

# VII. CONCLUSION

The Bayesian Hierarchical model provides a posterior distribution on weights and is suitable for generating weights to check the nature of the solution set. Moreover, generation of weights using the proposed Bayesian methodology can be used to develop a bona-fide Bayesian posterior distribution for the optima, thus properly and coherently quantifying the uncertainty about the optima. It has been shown that the proposed estimator outperforms the existing ones with respect to efficiency for small sample sizes. In practice, as it is difficult to have a large sample data, the effectiveness of the proposed method gains prominence. Bayesian determination of weights finds high applicability in cases where conducting a large scale survey is time consuming, difficult to implement as well as expensive. This proposed model is designed to adequately derive information from the collected data, rendering highly efficient estimators for small data sizes. This technique has been analyzed for error variances thereby quantifying the reliability of the estimates.

When applied in the domain of route optimization in discovering the most suitable parking lot, the proposed methodology have produced results which display close resemblance to the phenomenon observed in real life situations. This work relied on sound statistical techniques to improve the weights representing the relative importance of the possibly conflicting objective functions of the route optimization process rather than improving the solution set directly. It has also been observed that on an average the fitness values obtained under weights generated by the proposed methodology outperforms that obtained by frequentist approach. If implemented in reality, this would certainly ensure saving of time, energy and fuel, thus a greener world.